# UniGlyph: A Seven-Segment Script for Universal Language Representation

Bency Sherin, G. V.

Department of Electronics and Communication Engineering, St. Xavier's Catholic College of Engineering, Chunkankadai, Tamil Nadu 629003, India.

Abijesh Euphrine, A.

Deen Dayal Upadhyay Centre, Loyola College, Affiliated to University of Madras, Chennai, Tamil Nadu 600034, India.

Lenora Moreen, A.

Bell Matriculation Higher Secondary School, Palayamkottai, Tamil Nadu 627002, India.

Arun Jose, L.*

Department of Physics, St. Xavier's College, affiliated to Manonmaniyam Sundaranar University, Palayamkottai, Tamil Nadu 627002, India., arunjose@stxavierstn.edu.in

UniGlyph represents a groundbreaking venture into the realm of constructed languages (conlangs), aiming to establish a universal transliteration system using a script derived from seven-segment characters. Its core objective is to foster cross-language communication through the provision of a flexible and standardized script capable of accurately representing a broad spectrum of phonetic sounds. This innovative system addresses the imperfections of the International Phonetic Alphabet (IPA) and the limitations of traditional character sets by offering a compact and versatile method to represent phonetic diversity across languages. This paper delves into the intricate design of UniGlyph, delineating its script structure, phonetic mapping, and transliteration rules. The essence of UniGlyph lies in its unique design, characterized by a succinct character set augmented with markers for pitch and length variations. This streamlined script facilitates efficient representation of phonetic elements, enabling seamless transliteration from diverse languages into a cohesive script. The discussion also encompasses the challenges inherent in crafting a universal script and elucidates the guiding principles governing the transliteration process. By exploring various applications and use cases, from multilingual messaging to enhance artificial intelligence systems by providing a consistent and precise method for phonetic transcription, including future expansion possibilities such as the assignment of new scripts to previously unrepresented phonetic sounds and the potential for encoding animal phonetic communication. Each species could be assigned a distinct script followed by a specific number, allowing for the study and documentation of non-human vocal patterns. Through its innovative design and adaptability, UniGlyph presents a promising tool for bridging linguistic gaps and enhancing both human and non-human communication thereby improving language processing capabilities, the paper underscores the versatility and adaptability of UniGlyph. This system has the potential to revolutionize cross-language communication, offering an innovative solution to bridge linguistic divides. UniGlyph's robust transliteration system, coupled with its compact and versatile script design, presents a promising avenue for fostering a more interconnected world.

* Corresponding Author and all authors contributed equally to this research.

## 1 INTRODUCTION

Language has evolved significantly over millennia, reflecting the complex interplay between human communication needs and technological advancements [49]. From the intricate pictographs of ancient civilizations to the sophisticated scripts of today, writing systems have continually adapted to better capture and convey the nuances of spoken language [8, 41, 122]. The development of UniGlyph, a constructed language (conlang) featuring a seven-segment script for universal language representation, represents a modern effort to address the limitations of traditional writing systems and enhance cross-language communication [88].

### 1.1 Evolution of Writing Systems

The earliest forms of writing, such as cuneiform in Mesopotamia and hieroglyphs in Egypt, were characterized by their complexity and exclusivity. These scripts, comprised of intricate symbols and pictographs, were used primarily for administrative and ceremonial purposes and required specialized knowledge to interpret [39, 62, 113, 123]. This complexity limited literacy to a small group of scribes and scholars, creating a significant barrier to widespread communication and knowledge dissemination. As societies expanded and communication needs grew more diverse, the limitations of early writing systems became apparent [20]. The transition to more abstract and standardized scripts, such as the Phoenician alphabet around 1200 BCE, marked a significant advancement [41]. This alphabetic system simplified writing by using a limited set of symbols to represent sounds, making literacy more accessible and facilitating trade and cultural exchange [34, 40, 100, 102]. The Greek and Latin alphabets, derived from the Phoenician script, further standardized writing and played crucial roles in the development of Western languages [92]. The invention of the printing press by Johannes Gutenberg in the 15th century revolutionized writing by enabling the mass production of texts. This technological breakthrough significantly increased the availability of written materials, leading to higher literacy rates and the rapid spread of knowledge and ideas during the Renaissance [35, 92]. Despite these advancements, the complexity of accurately representing phonetic nuances in written form persisted [43].

### 1.2 Modern Challenges in Phonetic Representation

In contemporary times, the diversity of spoken languages and their phonetic complexities present significant challenges for transcription and digital communication [7, 10, 90, 95]. Traditional writing systems often struggle to capture the full range of phonetic sounds in languages, leading to inconsistencies and misunderstandings in transliteration [37, 93]. The International Phonetic Alphabet (IPA), developed to provide a standardized system for representing the sounds of spoken language, has been a critical tool in addressing these challenges [69, 71]. However, the IPA is not without its imperfections [14, 71, 105]. While it offers a comprehensive framework for phonetic transcription, its extensive set of symbols can be daunting for non-specialists to learn and use. Moreover, the IPA's typographic complexity poses challenges in digital representation, particularly on platforms with limited font support [13, 94]. These issues underscore the need for a simpler, more intuitive script that can effectively represent phonetic diversity across languages [17, 36, 78, 114].

### 1.3 The Need for a Simplified Script

The rise of digital technology and globalization has intensified the demand for efficient and versatile systems for cross-language communication [105]. In a world where individuals increasingly interact across linguistic boundaries, the ability to accurately represent and transmit diverse phonetic sounds is crucial [18, 21, 81, 103, 106, 119]. Traditional writing systems and the IPA, while valuable, often fall short in meeting these needs due to their complexity and limitations [60]. UniGlyph addresses these challenges by introducing a novel seven-segment script designed for universal phonetic representation [68]. Drawing inspiration from the geometric clarity and simplicity of seven-segment displays commonly used in digital devices, UniGlyph offers a compact and versatile method for capturing phonetic elements [12, 14, 15, 54, 88, 91]. This script, constructed from a combination of seven segments, provides a minimal yet sufficiently diverse character set that can be easily rendered on digital platforms.

### 1.4 Design Principles of UniGlyph

The design of UniGlyph focuses on creating a universal script that can accommodate the phonetic diversity of multiple languages. The script's structure includes representations for consonants, vowels, length variations, and pitch variations. By distinguishing between continuous and non-continuous consonants, UniGlyph captures the phonetic properties of sounds more accurately [99]. Continuous consonants, such as O {s} (s) and m {m} (m) (The script written inside the curly brackets '{ }' follows the ISO 15919 standard – Transliteration of Devanagari and related Indic scripts into Latin characters [1, 5, 6], while the text inside the normal brackets '( )' represents the International Phonetic Alphabet [124]), can be sustained without altering their phonetic quality, while non-continuous consonants, such as q {t) (t) and k {k} (k), are characterized by brief, explosive sounds [25, 63, 68, 77, 124, 125]. Vowels in UniGlyph are represented using distinct segment combinations that reflect their phonetic characteristics. The script includes mechanisms for indicating short and long variations of sounds, enabling precise representation of phonetic length. Additionally, pitch variations are incorporated to allow the transcription of tonal languages, where pitch plays a crucial role in distinguishing meaning [23, 27].

### 1.5 Addressing IPA Imperfections

UniGlyph's design addresses several key imperfections of the IPA. The IPA's extensive symbol set, while comprehensive, can be difficult to master and use, particularly for individuals without linguistic training [16, 42, 58, 71]. while useful for representing the sounds of spoken languages, has some limitations. It provides limited coverage for important linguistic elements like tone, stress, and prosody. Additionally, mastering the IPA requires significant effort due to the complexity and number of specialized symbols, making it challenging for casual users [52, 79]. There is also inconsistency in its application, with different linguists and publications sometimes using varying conventions. Technological challenges further complicate its use, as inputting and displaying IPA symbols can be cumbersome on platforms without native support [53, 64, 82]. Moreover, the IPA transcription often differs from a language's standard orthography, creating difficulties for language learners in connecting written and spoken forms. UniGlyph simplifies phonetic representation by using a smaller, more intuitive set of characters, making it accessible to a broader audience [98]. The geometric clarity of UniGlyph's seven-segment characters ensures they are well-suited for digital representation, overcoming the typographic limitations associated with the IPA [26, 72, 126]. Moreover, UniGlyph's adaptability allows it to accommodate new pronunciation characters as linguistic needs evolve. This flexibility ensures that UniGlyph remains relevant and useful in the face of ongoing linguistic developments, providing a robust framework for future enhancements [14]. This adaptability positions UniGlyph as a versatile tool for phonetic transcription in a rapidly changing linguistic landscape.

### 1.6 Conclusion

UniGlyph represents a significant step forward in the quest for a universal language representation system. By simplifying the transcription of phonetic sounds while allowing for expansion and flexibility, it provides a practical solution to the challenges posed by cross-language communication and linguistic diversity. The potential for integrating animal phonetic systems into this framework further expands its relevance, making it a tool that could benefit not just human communication but the broader understanding of communication as a biological phenomenon. As AI continues to shape the future of communication, a system like UniGlyph, which bridges linguistic gaps and accommodates a wide range of phonetic systems, will be crucial in creating a more interconnected and understanding world.

## 2 BACKGROUND AND MOTIVATION

The evolution of language and its representation has always been central to the development of human civilization. From the earliest forms of writing, such as cuneiform and hieroglyphics, to modern alphabets and phonetic scripts, the representation of speech sounds has been a challenge that societies have continuously sought to address. In ancient times, written communication was complex and accessible to only a few, as it required mastering intricate symbols, often numbering in the thousands. As languages and societies evolved, so did the need for simplified, yet comprehensive writing systems that could accurately represent the wide array of sounds produced in human speech. Modern scripts like the Roman, Cyrillic, and Devanagari alphabets have addressed this need to varying degrees, each tailored to specific linguistic families.

However, the increasing interconnectedness of the world in the 21st century has exposed the limitations of traditional scripts and phonetic systems. As global communication becomes more integral to commerce, science, education, and diplomacy, the demand for a universal script capable of representing diverse languages has become apparent. The International Phonetic Alphabet (IPA) was a groundbreaking attempt to create such a system, offering a standardized way to represent phonetic sounds across languages. Yet, despite its comprehensive design, the IPA remains a complex and difficult-to-master system, particularly for non-linguists. Furthermore, the IPA's rigid structure does not easily accommodate the dynamic nature of language evolution, nor does it adequately address the phonetic needs of tonal languages or dialects with continuous and non-continuous consonant distinctions [65].

These limitations have given rise to the development of UniGlyph, a seven-segment-based script designed to provide a more flexible and versatile solution for representing the phonetic diversity of human languages. UniGlyph builds upon the concept of using a compact character set, much like the digital displays used in calculators and electronic devices, to represent phonetic elements in a visually efficient way. The need for such a system has been driven by several factors, including the shortcomings of the IPA and traditional character sets in representing a wide range of phonetic sounds, particularly for tonal languages and languages with unique consonant structures [48].

### 2.1 The Challenge of Limited Character Sets

The evolution of human language has brought about an incredible diversity of spoken sounds, each carrying unique phonetic characteristics that can be challenging to capture in written form [2, 19, 21, 66, 70, 76, 79, 79, 81, 86, 89]. Traditional character sets, while functional, often fall short in representing the full range of phonetic nuances present in different languages [16, 18, 21, 97, 105, 107, 127]. The Roman alphabet, for instance, is widely used but lacks the flexibility to accommodate the vast phonetic diversity of global languages without additional diacritical marks or modifications [16, 31, 52, 74, 84, 112, 112, 119]. One of the significant challenges in linguistic representation is the need for a universal system that can accurately and consistently capture these phonetic elements. Existing systems like the International Phonetic Alphabet (IPA) aim to address this issue by providing a comprehensive set of symbols designed to represent all the sounds of spoken languages. However, the complexity and extensive nature of the IPA can make it inaccessible to non-specialists and challenging to implement in digital formats. This complexity often leads to inconsistencies and errors in transcription, hindering effective cross-language communication.

### 2.2 Why Seven-Segment Characters?

The idea of using seven-segment characters for language representation stems from their simplicity and widespread recognition in digital displays. Seven-segment displays, commonly found in digital clocks, calculators, and other electronic devices, use a combination of seven segments to form numeric and some alphabetic characters [11, 96, 104, 109]. This geometric simplicity ensures that the characters are easily recognizable and can be rendered across various digital platforms without losing clarity. By adopting a seven-segment script, UniGlyph leverages this simplicity to create a compact and versatile method for representing phonetic diversity. Each character in UniGlyph is constructed from a combination of seven segments, allowing for a limited but sufficient character set that can capture essential phonetic elements. This design not only simplifies the learning process but also ensures that the script is easily translatable to digital formats, addressing the typographic challenges faced by more complex systems like the IPA.

### 2.3 Complexity in Writing: Ancient and Modern Contexts

Throughout history, the complexity of writing systems has been a significant barrier to literacy and effective communication [59, 87]. In ancient times, scripts like cuneiform and hieroglyphs were intricate and required specialized knowledge to interpret [20, 28, 32, 50, 61, 85, 113, 128]. These scripts served administrative and ceremonial purposes but were inaccessible to the general populace, limiting their utility for widespread communication. As languages evolved, there was a clear need for simpler and more standardized scripts. The development of alphabets like the Phoenician, Greek, and Latin scripts marked a pivotal shift towards more accessible writing systems. These alphabets reduced the complexity of writing by using a limited set of symbols to represent sounds, paving the way for broader literacy and more efficient communication [8, 24, 30, 44, 51, 80, 129]. However, even these advancements could not fully capture the phonetic richness of spoken languages, leading to the continued evolution of writing systems. In the modern era, the advent of digital technology has further highlighted the need for simple and versatile scripts. Digital communication requires scripts that can be easily rendered on various devices and platforms. Traditional character sets, while functional in print, often struggle to maintain clarity and consistency in digital formats. This has driven the search for innovative solutions like UniGlyph, which can bridge the gap between phonetic accuracy and digital efficiency.

### 2.4 Imperfections in the International Phonetic Alphabet (IPA)

IPA is a standardized system for phonetic transcription that has been widely adopted by linguists and language educators. The IPA provides a comprehensive set of symbols designed to represent the sounds of all spoken languages, offering a valuable tool for phonetic analysis and language learning. However, the IPA's extensive symbol set and typographic complexity present significant challenges. Firstly, the IPA's complexity can be daunting for non-specialists to learn and use. The sheer number of symbols and their intricate designs require a high level of familiarity and expertise, making the IPA less accessible to the general public. This complexity can lead to inconsistencies in transcription and difficulties in accurately representing phonetic elements, especially in multilingual contexts. Secondly, the IPA's typographic nature poses challenges in digital representation. Many digital platforms and devices have limited support for IPA symbols, leading to rendering issues and loss of phonetic detail. These limitations hinder the effective use of the IPA in digital communication, where clarity and consistency are paramount. UniGlyph addresses these imperfections by offering a simplified and intuitive script that retains phonetic accuracy. The seven-segment design ensures that characters are easily recognizable and digitally translatable, overcoming the typographic challenges associated with the IPA. By providing a more accessible and versatile system, UniGlyph enhances the representation of phonetic diversity across languages.

### 2.5 Versatility for Future Adaptations

One of the key strengths of UniGlyph is its adaptability. While the initial character set is designed to capture a broad range of phonetic elements, the system is versatile enough to accommodate new pronunciation characters as linguistic needs evolve. This adaptability ensures that UniGlyph remains relevant and useful in the face of ongoing changes in language and phonetics. Linguistic evolution is a constant process, driven by cultural exchange, technological advancements, and changes in communication practices. New sounds and phonetic variations emerge, requiring updates to existing transcription systems. UniGlyph's modular design allows for the seamless integration of new characters and symbols, ensuring that the script can evolve alongside language. This adaptability is crucial for the long-term success of a universal transliteration system. By providing a flexible framework, UniGlyph can continue to meet the needs of diverse linguistic communities and contribute to more effective cross-language communication. This versatility positions UniGlyph as a forward-looking solution that can accommodate future developments in phonetics and linguistics.

### 2.6 Potential application in artificial intelligence

Another key motivation for UniGlyph's development is its potential application in artificial intelligence (AI) [67, 88, 101, 116, 118, 119]. In recent years, AI has made significant strides in areas such as speech recognition, natural language processing (NLP), and machine translation. However, these technologies often struggle with the variability and complexity of human language, particularly when dealing with multiple languages or dialects. The ability to accurately transcribe and analyze phonetic data is crucial for improving the performance of AI systems in these areas. UniGlyph offers a solution by providing a standardized, consistent method for representing phonetic sounds across different languages. This consistency can help reduce errors in AI-driven language processing, leading to more reliable and effective applications. For example, in speech recognition systems, UniGlyph can be used to improve the accuracy of phonetic transcription, ensuring that the AI correctly interprets and processes spoken input. Similarly, in machine translation, UniGlyph can facilitate more accurate transliteration of source texts, reducing the risk of mistranslation and improving the overall quality of the output. Furthermore, UniGlyph's design allows for future adaptability. As languages continue to evolve and new sounds emerge, the system can be expanded to include additional characters, ensuring that it remains relevant and useful in the face of linguistic change. This adaptability is particularly important in the context of AI, where the ability to quickly incorporate new linguistic data is essential for keeping pace with the rapid development of language technologies.

### 2.7 Expanding UniGlyph to Non-Human Communication

Another exciting frontier for UniGlyph lies in its potential application to the study and representation of animal phonetic systems. While the current focus of UniGlyph is on human languages, future research may extend its scope to include the structured sound systems used by certain animal species. Animal communication, particularly in species such as dolphins, birds, and primates, has been shown to exhibit complex patterns of vocalization that share similarities with human phonetics [121]. These sound systems, though not as linguistically structured as human languages, nonetheless follow recognizable patterns of pitch, duration, and tonal variation. UniGlyph's adaptability makes it well-suited to the task of representing these animal phonetic systems in a systematic and standardized way. As researchers in bioacoustics and animal communication continue to explore the intricacies of non-human vocalization, UniGlyph could offer a framework for transcribing and analyzing these sounds, much as the IPA does for human languages. By expanding its utility beyond human communication, UniGlyph could become a valuable tool in the fields of animal behavior research and interspecies communication, further broadening its impact.

### 2.8 Conclusion

The development of UniGlyph is motivated by the need for a simplified and versatile script that can accurately represent phonetic diversity across languages. Traditional writing systems and the IPA, while valuable, face significant challenges in terms of complexity, accessibility, and digital representation. UniGlyph addresses these issues by leveraging the geometric simplicity of seven-segment characters to create a compact and efficient phonetic script. By distinguishing between continuous and non-continuous consonants and incorporating mechanisms for pitch and length variations, UniGlyph ensures accurate phonetic representation. Its adaptability allows for the inclusion of new pronunciation characters as linguistic needs evolve, making it a versatile tool for phonetic transcription. In a world where cross-language communication is increasingly important, UniGlyph offers a promising solution for bridging linguistic gaps. Its innovative design and robust transliteration system enhance the representation of phonetic elements, contributing to a more interconnected and linguistically diverse global community. Its application in AI, in particular, highlights the system's potential to improve the accuracy and effectiveness of language processing technologies, making it a valuable tool for the future of communication.

## 3 SCRIPT DESIGN

UniGlyph is designed with simplicity and versatility in mind, making it an ideal foundation for a universal script applicable to a wide range of uses, including artificial intelligence (AI) systems. The goal is to create a universal transliteration system that accurately represents the phonetic diversity of languages. The script utilizes seven-segment characters, offering a unique and efficient approach through their geometric simplicity. This section explores the different components of the UniGlyph script, including the character set, consonants, vowels, variations in sound length and pitch, and other key considerations.

### 3.1 Character Set

The core of UniGlyph's design lies in its limited yet comprehensive character set. Each character is constructed from a combination of seven segments, similar to those used in digital displays. This geometric simplicity of the seven-segment model ensures that UniGlyph can be rendered on any device that supports basic digital display technology, from low-resolution screens to advanced AI-driven interfaces. This adaptability makes UniGlyph a versatile tool for both human and machine use, enabling seamless communication across different languages and platforms. The character set includes symbols for consonants, vowels, and markers for variations in sound length and pitch.

### 3.2 The use of seven-segment characters offers several advantages

- **Simplicity**: The limited number of segments ensures that characters remain simple and easy to learn.

- **Consistency**: The geometric design allows for consistent rendering on various devices and platforms.
- **Versatility**: Despite the limited number of segments, the character set is versatile enough to represent a wide range of phonetic sounds.

### 3.3 Consonants

UniGlyph includes a set of consonant characters designed to capture the essential phonetic features of consonants across different languages. The consonant characters are divided into two categories: continuous and non-continuous consonants.

- **Continuous Consonants**: These consonants, such as nasals and fricatives, involve a continuous airflow through the vocal tract [22, 130]. Examples include ꟼ {m} (m), ꟻ {n} (n), and ɢ {s} (s) (The script written inside the curly brackets '{ }' follows the ISO 15919 standard – Transliteration of Devanagari and related Indic scripts into Latin characters [1], while the text inside the normal brackets '( )' represents the International Phonetic Alphabet [124]). Continuous consonants in UniGlyph are represented by characters that include specific segments indicating continuous phonation.
- **Non-Continuous Consonants**: These consonants, such as plosives and stops, involve a complete closure and release of airflow. Examples include ᴘ {p} (p), Я {t} (t), and ┘ {k} (k). non-continuous consonants are represented by characters with different segment combinations to distinguish them from continuous consonants.

Each consonant character in UniGlyph is mapped to its corresponding sound in various languages, ensuring accurate phonetic representation. This mapping takes into account the International Phonetic Alphabet (IPA) and other standard transliteration systems to maintain consistency and accuracy. In the UniGlyph script, consonants are represented by specific configurations of the seven segments. Each configuration corresponds to a distinct consonant sound, with continuous consonants (those that can be sustained) being distinguished from non-continuous consonants (those that cannot be sustained) by their unique segment patterns. This distinction is crucial for accurate phonetic representation and is particularly relevant in AI applications, where precise differentiation between sounds can improve the accuracy of speech recognition and natural language processing (NLP) systems. By providing a clear and consistent method for representing consonants, UniGlyph enhances the ability of AI systems to process and interpret spoken language as shown in **Error! Reference source not found.**. This is especially important in multilingual contexts, where the accurate identification of consonant sounds can significantly impact the performance of AI-driven language models.

Table 1: Detailed mapping of Consonants in UniGlyph

| Language origin | Indic script | ISO 15919 transliteration | International phonetic alphabet | Continuous consonants | Proposed seven segment displays | Assigned keyboard character |
|---|---|---|---|---|---|---|
| Tamil | க் | k | k, g, x, ɣ, h, ŋ | No | ┘ | k |
| Tamil | ங் | ṅ | ŋ | Yes | ʟ | A |
| Tamil | ச் | c | t͡ʃ, d͡ʒ, ʃ, s, ʒ | No | Ꮯ | c |
| Tamil | ஞ் | ñ | ɲ | Yes | ɼ | E |
| Tamil | ட் | ṭ | ʈ, ɖ, ɽ | No | ə | x |
| Tamil | ண் | ṇ | ɳ | Yes | ʜ | C |
| Tamil | த் | t | t̪, d̪, ð | No | Я | q |
| Tamil | ந் | n | n | Yes | ꟻ | y |
| Tamil | ப் | p | p, b, β | No | ᴘ | p |
| Tamil | ம் | m | m | Yes | ꟼ | m |
| Tamil | ர் | r | ɾ | Yes | ƽ | s |
| Tamil | ல் | l | l | Yes | ʟ | l |
| Tamil | வ் | v | ʋ | No | h | r |

| Language origin | Indic script | ISO 15919 transliteration | International phonetic alphabet | Continuous consonants | Proposed seven segment displays | Assigned keyboard character |
|---|---|---|---|---|---|---|
| Tamil | ழ் | l̲ | ɻ | Yes | ŧ | w |
| Tamil | ள் | ḷ | ɭ | Yes | ʜ | I |
| Tamil | ற் | t | t, d | No | ь | N |
| Tamil | ன் | ṉ | n | Yes | ʔ | S |
| Tamil Grantha | ஜ் | j | d͡ʒ | No | ᒐ | j |
| Tamil Grantha | ஶ் | ś | ɕ, ʃ | Yes | ᒍ | H |
| Tamil Grantha | ஷ் | ṣ | ʂ | Yes | ⊣ | v |
| Tamil Grantha | ஸ் | s | s | Yes | ɢ | O |
| Tamil Grantha | ஹ் | h | h | No | ɦ | T |
| Devanagari | ग् | g | ɡ | No | ∩ | g |
| Devanagari | ड् | ḍ | ɖ | No | ʕ | R |
| Devanagari | द् | d | ð | No | ԁ | d |
| Devanagari | ब् | b | b | No | ⊢ | b |
| Devanagari | ज़् | z | z | Yes | ɿ | z |
| Devanagari | झ़् | zh | ʒ | Yes | ɜ | D |
| Devanagari | फ़् | f | f | Yes | ꜰ | f |

### 3.4 Vowels

Vowels in UniGlyph are similarly represented by unique seven-segment configurations, each corresponding to a specific vowel sound. The script includes provisions for both short and long vowel variations, ensuring that the nuances of vowel length are accurately captured. This is essential for maintaining the integrity of phonetic transcription, particularly in languages where vowel length can alter the meaning of words [3, 4, 9, 38]. The accurate representation of vowels is critical for AI applications such as text-to-speech (TTS) systems, where vowel length and pitch variations must be correctly interpreted to produce natural-sounding speech. UniGlyph's design allows for these variations to be encoded in a simple and consistent manner, reducing the potential for errors in AI-driven language processing as shown in **Error! Reference source not found.**.

Table 2: Detailed mapping of Vowels in UniGlyph

| Language origin | Indic script | ISO 15919 transliteration | International phonetic alphabet | Proposed seven segment displays | Assigned keyboard character |
|---|---|---|---|---|---|
| Russian | | | ɨ | ʁ | i |
| English | | | ɛ | ϱ | n |
| Tamil | அ | a | ʌ | ᴀ | a |
| Tamil | இ | i | i | ᴦ | e |
| Tamil | எ | e | e | Ɛ | t |
| Tamil | உ | u | u, ɯ | ᑌ | u |
| Tamil | ஒ | o | o | ٩ | o |

### 3.5 Short and Long Variations

UniGlyph incorporates markers for variations in sound length, distinguishing between short, long, very short, and very long sounds. These markers are essential for accurate phonetic representation, as the length of a sound can significantly affect its meaning in many languages. In AI-driven language models, the ability to recognize and process these variations can lead to more accurate speech recognition and synthesis, improving the overall quality of AI-based communication systems as shown in **Error! Reference source not found.**. For example, in languages like Japanese, where vowel length can change the meaning of a word, UniGlyph's markers for short and long variations can help AI systems correctly interpret and translate spoken input [57, 73, 110].

- **Short Sounds**: Represented by a base character followed by a specific marker indicating short duration.
- **Long Sounds**: Represented by the base character followed by a different marker indicating longer duration.
- **Very Short and Very Long Sounds**: Additional markers are used to indicate very short or very long durations, ensuring precise phonetic transcription.

Table 3: Detailed mapping of Short and Prolong Variations of Sound in UniGlyph

| Short and long variations | Proposed seven segment displays | Assigned keyboard character |
|---|---|---|
| Shot | _ | _ |
| Long | ¯ | ¯ |
| Example for Shot | A_ | A_ |
| Example for Very Shot | A__ | A__ |
| Example for Long | A¯ | A¯ |
| Example for Very Long | A¯¯ | A¯¯ |
| Example for Prolong width defined number | A¯4 | A¯4 |

### 3.6 Examples of sound length variations in UniGlyph

Short: a represented by a specific segment combination.

- **Long**: A¯ represented by the base segment combination followed by a length marker.
- **Very Short**: A_ represented by the base segment combination followed by a very short marker.
- **Very Long**: A__ represented by the base segment combination followed by a very long marker.

### 3.7 Pitch Variation

Pitch variation is another critical aspect of phonetic representation in UniGlyph. Pitch can convey different meanings and nuances in many languages, especially tonal languages [45, 55]. UniGlyph includes markers for different pitch levels, ensuring accurate transcription of pitch variations which are essential for conveying meaning in tonal languages such as Mandarin or Thai. The ability to encode pitch variations directly into the script is a significant advantage for AI applications, where the correct interpretation of pitch is critical for accurate speech recognition and synthesis as shown in **Error! Reference source not found.**. In AI-based translation and communication systems, pitch variations play a crucial role in ensuring that the intended meaning of a speaker is accurately conveyed. UniGlyph's design allows these pitch variations to be seamlessly integrated into the transliteration process, providing a robust framework for AI systems to process tonal languages effectively.

- **Normal Pitch**: Represented by the base character without any additional markers.
- **High Pitch**: Represented by the base character followed by a marker indicating high pitch.
- **Low Pitch**: Represented by the base character followed by a marker indicating low pitch.

- **Very High** and **Very Low Pitch**: Additional markers are used to indicate very high or very low pitch levels.

Table 4: Detailed mapping of Pitch Variations of Sound in UniGlyph

| Pitch variations | Proposed seven segment displays | Assigned keyboard character |
| --- | --- | --- |
| Very Very High Pitch |  | Z |
| Very High Pitch |  | Y |
| High Pitch |  | X |
| Normal Pitch |  | Q |
| Low Pitch |  | U |
| Very Low Pitch |  | V |
| Very Very Low Pitch | 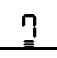 | W |

### 3.8 Examples of pitch variation in UniGlyph

- **Normal Pitch**: Q represented by a specific segment combination.
- **High Pitch**: X represented by the base segment combination followed by a high pitch marker.
- **Low Pitch**: U represented by the base segment combination followed by a low pitch marker.
- **Very High Pitch**: Y represented by the base segment combination followed by a very high pitch marker.
- **Very Low Pitch**: V represented by the base segment combination followed by a very low pitch marker.

## 4 ADDITIONAL CONSIDERATIONS

While UniGlyph presents significant advancements in the realm of phonetic transcription and language representation, there are several additional considerations that merit attention as we look toward its broader implementation and future development. These considerations involve challenges in standardization, the complexities of integrating UniGlyph into current linguistic and technological frameworks, potential ethical concerns, and the role of interdisciplinary research in ensuring the system's long-term success.

### 4.1 Standardization and Global Adoption

A major challenge for any new linguistic system is the process of achieving widespread standardization and acceptance. For UniGlyph to succeed as a universal script, it must be adopted by linguists, technologists, educators, and policymakers worldwide. This would require extensive collaboration across different sectors and regions to ensure that the system meets the needs of diverse linguistic communities. Unlike the International Phonetic Alphabet (IPA), which has had more than a century to establish itself, UniGlyph would need a coordinated effort to gain traction, especially among communities unfamiliar with its underlying principles. Standardization efforts must also consider the potential need for regulatory frameworks that oversee the integration of UniGlyph into educational materials, translation tools, and AI platforms. International linguistic bodies such as the International Phonetic Association, the United Nations Educational, Scientific and Cultural Organization (UNESCO), and global technology leaders may play pivotal roles in this process. The collaborative effort needed to integrate UniGlyph into these domains will be a critical step toward achieving its goal of becoming a truly universal script.

### 4.2 Integration with Existing Linguistic Systems

The practical integration of UniGlyph into existing linguistic systems raises important considerations regarding compatibility and co-existence with current writing systems and phonetic transcriptions. While UniGlyph offers a streamlined and compact system, its design must accommodate languages with complex phonetic structures, including those with unique tonal variations, intricate consonant clusters, and grammatical structures that differ significantly from English or other Indo-European languages. Additionally, successful implementation will depend on the system's ability to coexist with widely used scripts, such as Latin, Cyrillic, Arabic, and Chinese characters. These scripts are deeply ingrained in the cultural and historical identities of various communities, and any attempt to introduce a universal script must be sensitive to the linguistic heritage of these groups. Ensuring that UniGlyph can be used in conjunction with traditional writing systems, rather than replacing them, will be key to its acceptance.

### 4.3 Ethical Considerations in Language Standardization

The introduction of any universal system, particularly one related to language, raises important ethical considerations. Language is a key component of cultural identity, and the imposition of a standardized system could be viewed by some communities as a form of linguistic imperialism. While UniGlyph is designed to enhance communication and foster understanding across linguistic barriers, its implementation must be carried out in a way that respects the autonomy and linguistic diversity of all communities. The risk of marginalizing lesser-known languages or dialects by prioritizing certain phonetic structures over others must be carefully addressed in the development and promotion of UniGlyph. Moreover, the application of UniGlyph in AI-driven systems introduces concerns related to data privacy, bias, and the potential for misuse in areas such as surveillance or manipulation of information. As AI continues to evolve, ensuring that linguistic data is used ethically and responsibly will be paramount to maintaining public trust in technologies that rely on systems like UniGlyph.

### 4.4 Future Expansion: Non-Human Communication

As mentioned earlier, UniGlyph has the potential to extend beyond human phonetic systems into the realm of animal communication. Research in bioacoustics and animal behavior has revealed that many species, including cetaceans (dolphins and whales), birds, and primates, possess complex vocalization systems that may one day be transcribed using a version of UniGlyph. This expansion of the system could revolutionize how we study animal communication, potentially leading to new insights into the cognitive abilities and social structures of non-human species. However, this expansion will require further development of the UniGlyph script to accommodate the unique sound patterns and frequencies observed in animal vocalizations. Unlike human languages, animal sounds often involve frequencies and patterns beyond the typical range of human hearing or speech, which presents new challenges in terms of phonetic transcription. Nevertheless, this remains an exciting frontier for future research, and the successful adaptation of UniGlyph for animal communication could open up entirely new fields of study.

For human communication, the expansion will involve assigning new scripts to phonetic sounds that have not yet been represented. As language evolves and new phonetic variations emerge across different dialects and linguistic contexts, UniGlyph's adaptable framework allows for the seamless inclusion of these sounds. By assigning specific glyphs to newly identified phonetic elements, the system can evolve in tandem with linguistic changes, ensuring that it remains relevant and comprehensive in representing human speech.

For non-human communication, UniGlyph envisions an exciting opportunity to document and encode animal phonetics. Each species exhibits distinct vocal patterns and sounds that could be assigned unique scripts. This will involve assigning a new script for each species, followed by a distinct number to represent that particular species. For example, birds, dolphins, or other vocal animals could have their unique glyph systems, with numbers indicating species differentiation.

This expansion would not only create a comprehensive phonetic database but could also be an invaluable tool for studying animal communication and cross-species interaction.

### 4.5 Challenges in AI Integration

While UniGlyph offers promising applications for artificial intelligence, particularly in speech recognition and natural language processing, its integration into existing AI frameworks presents some technical challenges. Current AI models, particularly those based on machine learning and neural networks, are often trained using large datasets that rely on existing phonetic systems such as the IPA. Transitioning these models to UniGlyph would require retraining on new datasets that accurately represent phonetic sounds using the UniGlyph system. Furthermore, the success of AI applications in translation and speech recognition depends heavily on the quality and consistency of the data used to train these systems. Ensuring that UniGlyph can provide the necessary accuracy and comprehensiveness across different languages, dialects, and accents will be crucial to its effectiveness in AI applications. Collaborations between linguists, computer scientists, and AI developers will be essential to overcoming these challenges.

### 4.6 Interdisciplinary Collaboration and Research

The development of UniGlyph is an inherently interdisciplinary endeavor, requiring expertise from fields as diverse as linguistics, computer science, education, artificial intelligence, and cognitive psychology. This collaboration is essential to ensure that the system meets both linguistic and technological requirements, while also being user-friendly and accessible to a broad audience. Future research into the applications of UniGlyph should explore how it can be optimized for different linguistic contexts, as well as its potential for integration into educational systems and digital communication platforms. Additionally, cognitive studies on the learning and usage of UniGlyph could provide valuable insights into how the human brain processes new scripts and phonetic systems. Understanding the cognitive load associated with learning UniGlyph, particularly for users with no prior experience in phonetics, will be an important consideration in determining the system’s long-term viability.

### 4.7 Conclusion

In summary, while UniGlyph presents a groundbreaking solution for phonetic representation across languages, several additional considerations must be taken into account to ensure its successful implementation and future development. These include challenges related to standardization, ethical concerns, and technical integration, as well as exciting opportunities for expansion into non-human communication and AI-driven applications. By addressing these considerations through interdisciplinary collaboration and ongoing research, UniGlyph has the potential to revolutionize global communication and foster greater understanding across linguistic and cultural boundaries.

## 5 PHONETIC MAPPING AND TRANSLITERATION RULES

The core of UniGlyph lies in its ability to accurately map phonetic sounds to a simplified and universally recognizable script. This process, known as phonetic mapping, is crucial for ensuring that UniGlyph can be used to represent the diverse range of sounds found in the world's languages. Transliteration, the process of converting text from one script to another while preserving pronunciation, is guided by a set of rules designed to maintain phonetic accuracy across different languages.

## 5.1 Phonetic Mapping

Phonetic mapping in UniGlyph is based on the precise identification of sounds, or phonemes, in a language and their corresponding representation using the seven-segment script. Each segment configuration in UniGlyph corresponds to a specific sound, ensuring that the script can faithfully represent the phonetic structure of any language. The simplicity of the seven-segment design allows for a direct and efficient mapping process, reducing the potential for ambiguity in pronunciation. In the context of artificial intelligence, accurate phonetic mapping is critical for applications such as speech recognition and natural language processing (NLP). AI systems rely on precise phonetic representations to correctly interpret spoken language and convert it into text. By providing a clear and consistent method for representing phonemes, UniGlyph enhances the ability of AI systems to process a wide range of languages, improving the accuracy of speech-to-text conversion and other language processing tasks.

## 5.2 Transliteration Rules

The transliteration rules of UniGlyph are designed to ensure that text converted from one language to another retains its original pronunciation as closely as possible. These rules take into account the phonetic differences between languages and provide guidelines for how sounds should be represented in the UniGlyph script. This process involves identifying equivalent sounds in the target language and mapping them to the corresponding UniGlyph characters. For AI-driven applications, these transliteration rules are essential for enabling accurate cross-language communication. In machine translation systems, for example, the ability to correctly transliterate names, technical terms, and other phonetic elements is crucial for producing translations that are both accurate and understandable. UniGlyph's transliteration rules provide a standardized approach to this process, ensuring that AI systems can handle the complexities of transliteration with minimal errors.

Table 5: Examples of Translation from different languages to UniGlyph

| Language | Script in native language | English equivalent | IPA | UniGlyph |
|---|---|---|---|---|
| English | The sun is shining brightly today | The sun is shining brightly today | ðə sʌn ɪz ˈʃaɪnɪŋ ˈbraɪtli təˈdeɪ | [illegible] |
| Mandarin Chinese | 今天的阳光明媚 | Jīntiān de yángguāng míngmèi | t͡ɕɪn˥ tʰjɛn˥ jaŋ˥ kwaŋ˥ miŋ˧ meɪ˥˩ | [illegible] |
| Spanish | Hoy el sol brilla intensamente hoy | El sol brilla intensamente hoy | el ˈsol ˈbrilla inˈtensaˈmente oi | [illegible] |
| Arabic | االيوم الشمس تشرق الزاهية | Al-yom ash-shams tashraqo az-zāhiyah | alˈjoːm ʔaʃˈʃams taʃraqʊ ʔazˈzaːhijah | [illegible] |
| Portuguese | Hoje o sol está brilhando intensamente hoje | Hoje o sol está brilhando intensamente hoje | ˈoʒi o ˈsɔw ˈesta brilˈjandu inˈtẽˈsamente ˈoʒi | [illegible] |
| French | Aujourd'hui, le soleil brille intensément aujourd'hui | Aujourd'hui, le soleil brille intensément aujourd'hui | oʒuʁˈdui lə soˈlɛj bril intensəmɑ̃ oʒuʁˈdui | [illegible] |
| Japanese | 今日、太陽は今日はとても明るく輝いています | Kyōjitsu, taiyō wa kyō wa totemo akiraka ni hikatte imasu | kʲoːdzit͡su taijoː wa kʲoː wa totemo akaraka ni hikatte imasu | [illegible] |
| Tamil | இன்று சூரியன் மிகவும் பிரகாசமாக பிரகாசிக்கிறது | Indru sūriyan mikavum pirakāśamakāga pirakāśikkiradu | indu suːrijan mikavum piɾakaːsamaːgaka piɾakaːsikkiɾadu | [illegible] |

| Language | Script in native language | English equivalent | IPA | UniGlyph |
|---|---|---|---|---|
| German | Heute scheint die Sonne heute sehr hell | Heute scheint die Sonne heute sehr hell | ˈhɔʏtə ˈʃaɪnt diː ˈzɔnə ˈhɔʏtə zeːr ˈhɛl | ĥAՐƏ ⅎAʔƏ ƏՐ ɾ9ʔA ĥAՐƏ ɾEƼ_ ĥEL |
| Hindi | आज सूरज बहुत चमक रहा है | Aaj sooraj bahut chamak rahā hai | aːdz suːrya bahut t͡ʃəmək raɦa hɛː | A¯Ƅ CU¯ƼAƄ ⱶĥUA CᴘAJ ƼAĥ ĥƎ |
| Bengali | সূর্য আজ খুব উজ্জ্বল | Sūrya āja khuba ujjbala | ˈsuːrjɔ ˈaːdʒ ˈkhub ˈuddʒɔl | ꟼU¯ƼƄA A¯Ƅ JUⱶUƼ ƄA¯L |
| Russian | Сегодня ярко светит солнце | Segodnya yarko svetit solntse | sʲevɐˈdʲnə ˈjarkə ˈsvʲetʲit ˈsolnt͡sɨ | GE9¯ƏɈe ՐAƼJA GhECՐA GAʔGe |
| Hebrew | השמש זורחת בבהירות היום | HaShamesh zoret be'behirut hayom | haʃaˈmeʃ zoˈret bevihiˈrut haˈjom | ĥAⅎᴘEⅎ ɾAhAĥAA ⱶՐƼeĥՐU AAՐ9L |
| Greek | ἥλιος λάμπει φωτεινῶς σήμερον | Hhēlios lampei phōteinōs sēmeron | ˈheːlios ˈlampi fotiˈnos ˈseːmeron | ĥՐՐAG LeⱶՐ FAAՐʔA¯G GՐᴘeƼA¯ʔ |
| Latin | Sol splendet clare hodie | Sol splendēt clārē hodiē | sɔl ˈsplɛndɛt ˈklare ˈhodie | G9L GPHEꟼƏEƏ JeƼ ĥA¯dE |
| Old Persian | [illegible] | Huvašra citiya hya haruva duvarš tiy | huvaʃra citija hja haruva duvarʃ tiy | ĥUhAⅎƼA GՐ¯ƏՐՐA ĥՐA ĥA¯Ƽ9A ʕ9hAⅎ ƏՐ |
| Dutch | De zon schijnt vandaag fel | De zon schijnt vandaag fel | də zɔn ʃxɛint fɑndaːx fɛl | de ɾ9ʔ GJEՐʔƏ hAdA¯Ո ULUP |
| Korean | 오늘 태양이 밝게 빛난다 | Oneul taeyang-i balkge bichnanda | onɯl tʰɛjaŋi palkːe pit͡ɕʰnanda | UɈEL AEՐAʔLՐ ⱶALJE PՐʔAꟼʕA |
| Swedish | Skiner solen klart idag | Skiner solen klart idag | ˈʃiːnər ˈsuːlən ˈklɑːʈ ɪˈdɑː | G9LEʔ ĥULʔEL GƏALJ ՐdA |
| Ancient Egyptian | [illegible] | Ra hrw šsp m dšr.w ḥr ibd | ra ħrw ʃsp m dʃrw ħr ibd | ƼA ĥEƼU ⅎEGP ᴘ dAⅎAƼU ĥAƼ ՐPA |
| Aramaic | ܫܡܫܐ ܕܫܡܫܐ ܠܩܕܫܐ ܝܘܡܐ | Shəmshā dshəmshā l'qādshā yawmā | ʃəmʃa dəʃəmʃa lqaːdʃa jawmaː | ⅎEᴘⅎA ʕEⅎEᴘⅎA LՈAƏⅎA ՐAhᴘA |
| Italian | il sole splende luminosamente oggi | il sole splende luminosamente oggi | il ˈsole ˈsplɛnde luˈminozamente ˈɔddʒi | L G9LE GPHEꟼʕE LUᴘՐʔ9 GAᴘEꟼAE AUՐ |
| Sumerian | [illegible] | Utu-da ud-lu-un-na-da-lil₂-ḫa | utu da ud luna da lil ħa | UdU dA Ud LUʔA dA LՐL ĥA |
| Akkadian | [illegible] | Šamaš ērib issak marduk zi' gub ma | ʃamaʃ eːrib issak marːduːk ziː guːb ma | JᴘAh E¯ƼՐⱶ ՐⅎⅎAJ ᴘAƼʕUJ GՐ ՈUⱶ ᴘA |

### 5.3 Handling Diverse Phonetic Elements

UniGlyph is designed to accommodate the wide variety of phonetic elements found in different languages, including sounds that may not have direct equivalents in other scripts. This versatility is achieved through the use of additional markers and modifiers that can be applied to the basic seven-segment characters to represent pitch, length, and other phonetic variations. These features allow UniGlyph to capture the full range of sounds in any language, ensuring that the script can be used for accurate phonetic transcription and transliteration as shown in **Error! Reference source not found.**. The inclusion of these markers is particularly important for AI applications that involve speech synthesis and recognition. In tonal languages, for example, pitch variation can change the meaning of a word, making it essential for AI systems to correctly identify and reproduce these variations. UniGlyph's ability to encode pitch and length variations directly into the script allows AI systems to process these elements more accurately, improving the overall performance of speech-based applications.

### 5.4 Integration with AI Systems

UniGlyph's design is inherently suited for integration with AI technologies, particularly in the areas of speech recognition, natural language processing, and machine translation. The script's simplicity and consistency make it an ideal candidate for use in AI systems that require efficient and accurate phonetic representation. By providing a standardized method for phonetic mapping and transliteration, UniGlyph can help reduce the complexity of language processing tasks in AI, leading to more reliable and effective language-based applications. For example, in automatic speech recognition (ASR) systems, UniGlyph can be used as an intermediate representation of spoken input, allowing the system to more accurately identify and process phonetic elements. This can lead to improved recognition accuracy, particularly in multilingual environments where traditional character sets may struggle to capture the nuances of different languages. Similarly, in machine translation systems, UniGlyph can provide a consistent framework for transliterating names, technical terms, and other non-translatable elements, enhancing the quality and coherence of the translated output.

### 5.5 Future Directions

As AI technologies continue to evolve, the need for flexible and adaptable language representation systems will only grow. UniGlyph's design allows for the easy addition of new characters and markers as linguistic needs change, ensuring that the script remains relevant and useful over time. This adaptability is particularly important for AI applications, where the ability to quickly incorporate new phonetic data can significantly impact system performance. In conclusion, UniGlyph's approach to phonetic mapping and transliteration offers a robust framework for representing the phonetic diversity of the world's languages. Its integration with AI technologies holds significant potential for improving the accuracy and efficiency of language processing tasks, making it a valuable tool for both human and machine communication in an increasingly interconnected world.

## 6 APPLICATION AND USE CASES

The versatility and adaptability of UniGlyph make it a powerful tool for a wide range of applications across various domains. Its potential extends beyond traditional linguistic contexts, offering innovative solutions for cross-language communication, education, symbolic representation, multicultural interactions, and creative endeavors. Moreover, its integration with artificial intelligence (AI) technologies opens up new possibilities for enhancing language processing, translation, and communication systems.

### 6.1 Cross-Language Communication

One of the primary applications of UniGlyph is in facilitating cross-language communication. By providing a universal script that can represent the phonetic structure of any language, UniGlyph enables speakers of different languages to communicate more effectively. This is particularly useful in multilingual regions where a common script can bridge language barriers, allowing for clearer and more accurate communication between diverse language groups. In AI-driven communication tools, such as chatbots and virtual assistants, UniGlyph can be employed to improve the accuracy and consistency of cross-language interactions. These AI systems can use UniGlyph as an intermediate script to transliterate and process inputs from multiple languages, ensuring that the intended meaning is preserved and understood across different linguistic contexts. This capability is especially valuable in globalized environments where effective communication across languages is essential.

### 6.2 Educational Contexts

UniGlyph also has significant potential in educational settings, particularly in the teaching of phonetics, linguistics, and language learning. Its simplified and consistent script can be used as a teaching aid to help students understand the phonetic elements of different languages. By providing a clear and unified representation of sounds, UniGlyph can make it easier for learners to grasp the nuances of pronunciation and phonetic variation. In AI-powered educational tools, such as language learning apps and pronunciation training software, UniGlyph can serve as a foundational script for teaching phonetics. AI systems can leverage UniGlyph to provide personalized feedback on pronunciation, helping learners to achieve more accurate and natural speech. The use of UniGlyph in these contexts can enhance the effectiveness of language education by providing a consistent and accessible framework for understanding phonetic principles.

### 6.3 Symbolic Representation

Beyond its use in language, UniGlyph can also be employed as a symbolic representation system for encoding phonetic information in a compact and visually distinct manner. This makes it suitable for use in various symbolic applications, such as shorthand writing, phonetic transcription, and even artistic expression. The ability to represent complex phonetic data with a limited set of symbols allows UniGlyph to function as a versatile tool for symbolic communication. AI systems that require efficient and accurate encoding of phonetic data can benefit from integrating UniGlyph into their symbolic processing algorithms. For example, in speech recognition systems, UniGlyph can be used to encode and analyze phonetic features, enabling more efficient and accurate processing of spoken language. This symbolic representation capability also extends to areas such as text-to-speech synthesis and linguistic data analysis, where precise phonetic encoding is critical for achieving high-quality results.

### 6.4 Multicultural Contexts

In multicultural and multilingual environments, UniGlyph offers a standardized script that can be used to represent the phonetic elements of multiple languages. This is particularly useful in contexts where individuals from different linguistic backgrounds need to communicate or collaborate. By providing a common script that is both neutral and universally applicable, UniGlyph can help to reduce language-related misunderstandings and foster more inclusive interactions. AI applications in multicultural contexts can leverage UniGlyph to enhance communication and collaboration between individuals from diverse linguistic backgrounds. For instance, AI-powered translation tools can use UniGlyph to transliterate and process multilingual inputs, ensuring that the original phonetic intent is preserved in the translated output. This can lead to more accurate and culturally sensitive translations, improving the quality of cross-cultural communication in various settings.

### 6.5 Creative Applications

The unique design of UniGlyph also lends itself to creative applications, where its distinct visual style can be used to explore new forms of artistic expression. Whether in graphic design, typography, or digital art, UniGlyph's seven-segment characters offer a visually striking and adaptable medium for creative experimentation. Artists and designers can use UniGlyph to create works that play with the boundaries between language, symbol, and visual art. In AI-driven creative tools, UniGlyph can be integrated as a font or script option, allowing users to explore new aesthetic possibilities in their digital creations. For example, AI-powered UniGlyph can be used to generate dynamic, real-time visualizations of speech or text, turning linguistic data into an artistic experience. The script's versatility and adaptability make it an ideal candidate for creative projects that seek to blend linguistic and visual elements in innovative ways.

### 6.6 AI-Powered Language Processing

UniGlyph's integration with artificial intelligence opens up new possibilities for enhancing language processing tasks. AI systems that involve speech recognition, natural language processing, and machine translation can benefit from UniGlyph's compact and consistent phonetic representation. By providing a standardized script for encoding phonetic elements, UniGlyph can improve the accuracy and efficiency of these AI-driven processes. In particular, AI-powered speech recognition systems can use UniGlyph as an intermediate representation of spoken input, enabling more precise identification and processing of phonetic features. This can lead to improved recognition accuracy, especially in multilingual environments where traditional character sets may struggle to capture the nuances of different languages. Additionally, in machine translation, UniGlyph can serve as a framework for transliterating names, technical terms, and other phonetic elements, enhancing the quality and coherence of the translated output.

Encode animal phonetic languages

Looking ahead, one of the most exciting future applications of UniGlyph lies in its potential to encode animal phonetic languages. As scientific research continues to explore the communication patterns of non-human species, UniGlyph could be adapted to represent the phonetic elements of animal calls, sounds, and vocalizations. This expansion of the system could contribute to fields like bioacoustics, zoology, and even AI, where understanding and interpreting animal communication is becoming a growing area of study. By providing a script for animal sounds, UniGlyph could support researchers in categorizing and analyzing vocalizations, ultimately enhancing our understanding of cross-species communication [29, 33, 46, 47, 56, 75, 83, 108, 111, 115, 117, 120].

### 6.7 Future Directions

The applications and use cases of UniGlyph demonstrate its potential to transform various aspects of language representation and communication. As AI technologies continue to advance, the integration of UniGlyph into AI systems holds significant promise for improving the accuracy and effectiveness of language processing tasks. The script's adaptability and versatility ensure that it can meet the evolving needs of linguistic research, education, and cross-language communication in an increasingly interconnected world. In conclusion, UniGlyph's innovative script design and robust transliteration system offer a flexible and powerful solution for representing phonetic diversity across languages. Its potential applications span a wide range of fields, from education and communication to creative expression and AI-driven language processing. As the world becomes more globalized and interconnected, UniGlyph's contributions to cross-language communication and linguistic research will continue to grow, making it a valuable tool for the future.

## 7 CONCLUSION

UniGlyph represents a significant advancement in the development of a universal transliteration system that bridges the gaps between different languages. Its design, rooted in a seven-segment display script, offers a compact, versatile, and efficient method for representing the phonetic diversity of languages worldwide. The system's ability to address the imperfections of the International Phonetic Alphabet (IPA) and the limitations of traditional character sets is a testament to its innovative approach to phonetic representation. The adaptability of UniGlyph is one of its most remarkable features. By distinguishing between continuous and non-continuous consonants, and incorporating mechanisms for pitch and duration variations, UniGlyph ensures accurate and nuanced phonetic representation. Furthermore, the script is designed to be future-proof, with the ability to accommodate new pronunciation characters as linguistic needs evolve. UniGlyph's potential applications are vast and varied. In cross-language communication, it serves as a universal script that facilitates clearer and more accurate interactions between speakers of different languages. In educational contexts, UniGlyph offers a consistent and accessible framework for teaching phonetics, helping learners understand the nuances of pronunciation across languages. Its use in symbolic representation and multicultural contexts further demonstrates its versatility and adaptability. One of the most exciting prospects for UniGlyph lies in its integration with artificial intelligence. As AI continues to play a central role in language processing, communication, and translation, UniGlyph offers a standardized script that can enhance the accuracy and efficiency of these AI-driven processes. Whether in speech recognition, natural language processing, or machine translation, UniGlyph provides a robust framework for encoding and analyzing phonetic data, leading to improved outcomes in multilingual environments. Additionally, future expansions of UniGlyph may involve integrating non-human communication systems, particularly animal phonetics, into the script. As research into animal communication continues to grow, UniGlyph could evolve to represent these phonetic patterns, allowing for a broader understanding of cross-species communication. The ability to encode animal sounds and calls would not only aid in scientific research but could also open new avenues for AI to interpret and interact with the natural world. The creative possibilities of UniGlyph also extend to the realms of art and design, where its distinct visual style can be used to explore new forms of artistic expression. AI-powered design tools can leverage UniGlyph to generate visually unique representations of phonetic data, further pushing the boundaries of creativity and innovation. In conclusion, UniGlyph stands out as a pioneering system that addresses the complex challenges of cross-language communication, phonetic representation, and language education. Its innovative script design, combined with its adaptability and potential for AI integration, positions UniGlyph as a valuable tool for the future. As the world becomes increasingly interconnected, UniGlyph's contributions to bridging linguistic divides and enhancing communication will continue to grow. With future expansions that may include the representation of animal phonetic languages, UniGlyph could even contribute to the broader understanding of communication beyond the human sphere, making it an indispensable asset in both language and technology.

## ACKNOWLEDGMENTS

We would like to express our sincere gratitude to Jeseentha V. for her invaluable assistance in helping with the pronunciation aspects of this project. Her expertise greatly contributed to the development and refinement of the phonetic elements of UniGlyph. No funding was received for this research.

## DECLARATION OF INTERESTS

The authors declare that they have no known competing financial interests or personal relationships that could have appeared to influence the work reported in this paper.

## DECLARATION OF GENERATIVE AI AND AI-ASSISTED TECHNOLOGIES IN THE WRITING PROCESS

During the preparation of this work, the author(s) used ChatGPT in order to generate initial drafts, refine language and brainstorm ideas. After using this tool, the author(s) reviewed and edited the content as needed and take(s) full responsibility for the content of the publication.